# A Binary Classification Framework for Two-Stage Multiple Kernel Learning

**Abhishek Kumar** ABHISHEK@CS.UMD.EDU
Department of Computer Science, University of Maryland, College Park, MD 20742, USA

**Alexandru Niculescu-Mizil** ALEX@NEC-LABS.COM
**Koray Kavukcoglu** KORAY@NEC-LABS.COM
NEC Laboratories America, Princeton, NJ 08536, USA

**Hal Daumé** HAL@UMIACS.UMD.EDU
Department of Computer Science, University of Maryland, College Park, MD 20742, USA

## Abstract

With the advent of kernel methods, automating the task of specifying a suitable kernel has become increasingly important. In this context, the Multiple Kernel Learning (MKL) problem of finding a combination of prespecified base kernels that is suitable for the task at hand has received significant attention from researchers. In this paper we show that Multiple Kernel Learning can be framed as a standard binary classification problem with additional constraints that ensure the positive definiteness of the learned kernel. Framing MKL in this way has the distinct advantage that it makes it easy to leverage the extensive research in binary classification to develop better performing and more scalable MKL algorithms that are conceptually simpler, and, arguably, more accessible to practitioners. Experiments on nine data sets from different domains show that, despite its simplicity, the proposed technique compares favorably with current leading MKL approaches.

## 1. Introduction

Kernel methods such as support vector machines (SVM) (Cortes & Vapnik, 1995), kernel ridge regression, or kernel PCA (Smola & Muller, 1999), use a positive semi-definite (PSD) *kernel* to implicitly map the instances from the original instance space to a feature space where the standard linear algorithm is applied. The main drawback of kernel methods is that they require the user to specify a single suitable kernel in the first place, which is often critical to the method's success, but is usually a hard task even when the user has a good familiarity with the problem domain. To ease this burden, significant attention has been given the problem of automatically learning the kernel. The majority of the previous work in this area has focused on the Multiple Kernel Learning (MKL) setting, where the user is only tasked with specifying a set of base kernels, and the learning algorithm is in charge of finding a combination of these base kernels that is appropriate for the problem at hand.

There have been two main lines of work in this direction. The first one learns both the the weights of the kernel combination and the parameters of the classifier by solving a single joint optimization problem. This *one-stage* approach was first proposed by (Lanckriet et al., 2004) and has since received significant attention (Rakotomamonjy et al., 2007; Sonnenburg et al., 2006; Cortes et al., 2010a; Kloft et al., 2011; Bach, 2008; Zien & Ong, 2007; Cortes et al., 2009; Sindhwani & Lozano, 2011).

The second line of work in kernel learning follows a two-stage approach: first learn a "good" combination of base kernels using the training data, then use the learned kernel with a standard kernel method such as SVM or kernel ridge regression to obtain a classifier/regressor. This approach has been initially proposed in (Cristianini et al., 2001) and (Kandola et al., 2002), and recently revisited by (Cortes et al., 2010b). The two-stage leaning approaches so far have been

Appearing in *Proceedings of the $29^{th}$ International Conference on Machine Learning*, Edinburgh, Scotland, UK, 2012.

based on the notion of *target alignment*. Intuitively, target alignment, is a measure of similarity (agreement) between a kernel and the *target kernel*, which is derived from the training labels, and represents the optimal kernel for the training sample.

In this paper we introduce TS-MKL, a general approach to Two-Stage Multiple Kernel Learning that encompasses the previous work based on target alignment as special cases. We formulate the kernel learning problem as a standard linear classification problem in a new instance space. In this space, any linear classifier with weights $\boldsymbol{\mu}$ directly corresponds to a linear combination of base kernels with weights $\boldsymbol{\mu}$. To avoid confusions, we will denote this new instance space as the *K-space*, and a classifier in the *K-space* as a *K-classifier* throughout the paper. Thus the problem of finding a "good" kernel combination reduces to finding a "good" linear classifier in the K-space, a very familiar problem. One big advantage of this approach is that one can easily adapt techniques from binary classification to solve the MKL problem. For instance, one can use familiar and well understood max-margin methods to obtain better performing MKL algorithms, or take advantage of the recent advances in large scale learning to scale up and/or parallelize the MKL implementations. For the results presented in this paper we learn K-classifiers (and hence kernels) by training $L_2$ regularized linear SVMs with positive weights using the stochastic projected sub-gradient descent method from Pegasos (Shalev-Shwartz et al., 2007).

On the theoretical side, we prove a finite sample generalization bound for the original classification task in terms of the expected hinge loss and the margin of a K-classifier in the K-space. This justifies our approach of training a K-classifier that has low hinge loss and high margin in the K-space in order to learn a good kernel for the original classification problem. To the best of our knowledge, this result represent the first finite sample bound for two-stage kernel learning, improving on previous bounds that were only asymptotic. We also give a concentration bound for the expected hinge loss of a K-classifier.

On the empirical side, we run a comprehensive evaluation on two object recognition datasets (Caltech 101 and 256), three bioinformatics datasets (Psort+, Psort-, Plant) and four UCI datasets. On all these datasets our method performs better than, or the same as target alignment, showing that choosing a better K-classifier is beneficial. Our method also fares well against one-stage multiple kernel learning approaches significantly outperforming them on Caltech-256 and being essentially tied on the others.

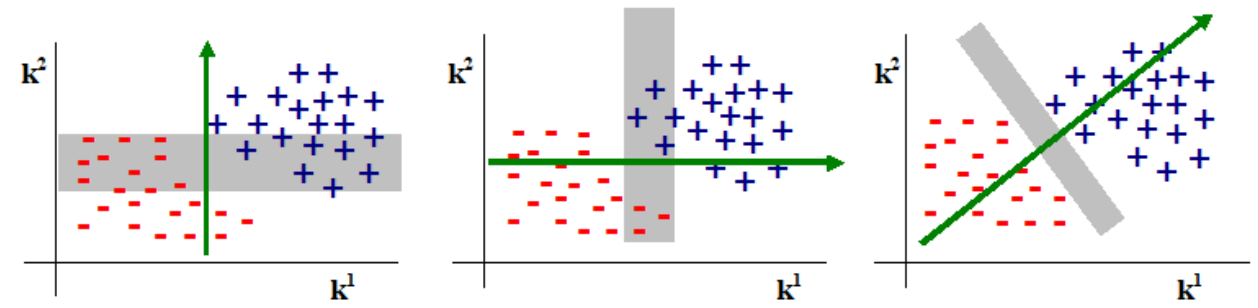

*Figure 1.* The K-space for two base kernels ($p = 2$). Points represent positive and negative K-examples $z_{xx'}$. The coordinates are the values of $K_1(x, x')$ and $K_2(x, x')$.

## 2. Method

We consider a classification problem where instances $(x, y)$ are drawn from a distribution $P$ over $\mathcal{X} \times \mathcal{Y}$, with $\mathcal{Y}$ a finite discrete set of labels. We assume that we have access to $p$ positive semi-definite (PSD) base kernel functions $K_1, \cdots, K_p$ with $K_i : \mathcal{X} \times \mathcal{X} \to \mathbb{R}$. Our goal is to learn a combination of these kernel functions that is itself a positive semi-definite function and is a "good" kernel for the classification task at hand. To achieve this, we define a new *binary* classification problem over a new instance space $\{(z_{xx'}, t_{yy'}) | ((x, y), (x', y')) \sim P \times P\} \subset \mathbb{R}^p \times \{\pm 1\}$ where

$$\begin{aligned} z_{xx'} &= (K_1(x, x'), \cdots, K_p(x, x')) \\ t_{yy'} &= 2 \cdot \mathbf{1}\{y = y'\} - 1 \end{aligned} \tag{1}$$

We will call this space the *K-space*, and call $z_{xx'}$ a *K-example* or *K-instance* and $t_{yy'}$ a *K-label*. Any function $h : \mathbb{R}^p \to \mathbb{R}$ in this space induces a similarity function $\tilde{K}_h$ between instances in the original space:

$$\tilde{K}_h(x, x') = h(z_{xx'}) = h(K_1(x, x'), \cdots, K_p(x, x'))$$

If $\tilde{K}_h$ is also positive semi-definite, hence a valid kernel, we say that $h$ is a *K-classifier*. For example, all linear functions with positive coefficients (i.e. $h_\mu(z_{xx'}) = \mu \cdot z_{xx'}$ with $\mu \geq 0$) are K-classifiers with the induced kernels $\tilde{K}_\mu$ being linear combinations of the $p$ base kernels. Figure 1 shows a toy example for the case of two base kernels. Each point in the figure is a labeled K-example $(z_{xx'}, t_{yy'})$ corresponding to a pair $(x, y), (x', y')$ of original instances. Note that the figure is drawn in K-space, not in input space. For a linear K-classifier $h_\mu$, the value of its induced kernel for a parir of original instances, $\tilde{K}_\mu(x, x')$, is the projection of the corresponding K-example $z_{xx'}$ on the vector $\mu$ (represented by the green line). The left and center sub-figures show the cases where $\mu$ is $(0, 1)$ and $(1, 0)$ respecively. In both cases the induced kernel combination is suboptimal. The linear combination in the right sub-figure corresponds to $\mu = (1, 1)$ and is a good combination because the kernel values of pairs of instances in the same class are separated from the kernel values of pairs of instances in different classes.

The key insight behind our method is that, if a K-classifier $h$ is a good classifier in the K-space, then the induced kernel $\tilde{K}_h(x, x') = h(z_{xx'})$ will likely be positive when $x$ and $x'$ belong to the same class and negative otherwise. This makes $\tilde{K}_h$ a good kernel for the original classification task. This intuition is made more precise in Section 3 where we provide a generalization bound that shows that a K-classifier that separates the positive and negative K-examples with high margin will indeed induce a kernel that allows learning a good classifier for the original task. Note that having a good K-classifier is a sufficient condition, not a necessary one. There can very well exist combinations of base kernels that do not correspond to a good K-classifier, but are good kernels nevertheless. Unlike one-stage kernel learning approaches, our method will not be able to find such combinations and it might miss on some good kernels. The results in Section 4, however, show that this does not seem to be the case in practice, as we consistently matched or exceeded the performance of one-stage MKL.

Thus the problem of learning a good kernel can be reduced to the problem of learning a good K-classifier in the newly defined K-space: given a training sample $(x_i, y_i)_{i=1}^n$ for the original classification task, construct a K-training set $(z_{ij}, t_{ij})_{1\le i\le j\le n}$ and learn a K-classifier $h$ from this sample. Any learning algorithm can be used for learning $h$ provided that the induced kernel can be guaranteed to be a valid PSD kernel[1].

In line with the majority of the MKL work, in this paper we focus on learning linear K-classifiers, and hence linear combinations of base kernels. The results in Section 3 suggest that it is desirable to have a maximum margin K-classifier, thus we use $L_2$ regularized linear SVM to learn the K-classifier, and ensure that the induced kernel is PSD by constraining the weights to be positive. One could, however, use a sparsity promoting regularizer (e.g., $L_1$ penalty) if a sparse combination of kernels is desired.

The optimization problem for learning the kernel weights $\boldsymbol{\mu}$ is thus given by

$$\min_{\boldsymbol{\mu}\ge 0} \frac{\lambda}{2}||\boldsymbol{\mu}||^2 + \frac{1}{\binom{n}{2}+n} \sum_{1\le i\le j\le n} [1 - t_{ij}\boldsymbol{\mu}\cdot \mathbf{z}_{ij}]_+ \quad (2)$$

where $[1-s]_+ = \max\{0, 1-s\}$ is the hinge loss.

To optimize this objective we use the stochastic projected sub-gradient descent implemented in Pegasos (Shalev-Shwartz et al., 2007), with an additional projection to the non-negative constraint set after every gradient step. Using a stochastic optimization method allows us to scale very well despite the quadratic number of K-examples: computation time is not directly dependent on the number of instances, linear in the number of base kernels, and independent of the number of classes. If needed, memory usage can be reduced through streaming techniques or on the fly construction of the K-examples.

[1]One could drop the PSD requirement and use any classifier, even a non-linear one, to obtain a *similarity function* rather than a proper kernel. The theory of learning with similarity functions (Balcan & Blum, 2006) can be then applied to learn a classifier for the original task. Generalization bounds similar to the ones in Section 3 would also hold for this case.

### 2.1. Connection to Target Alignment

Previous two-stage kernel learning approaches (Cristianini et al., 2001; Cortes et al., 2010b) learn a non-negative linear combination of base kernels that maximizes the *alignment* with the target kernel $K^{(t)}(\mathbf{x}_i, \mathbf{x}_j) = y_i y_j$ on the training set. This is achieved by solving the optimization problem

$$\max_{\boldsymbol{\mu}\ge 0} \frac{\langle \sum_{l=1}^p \mu_l \mathbf{K}_l, \mathbf{K}^{(t)}\rangle}{|| \sum_{l=1}^p \mu_l \mathbf{K}_l||_F}, \quad \text{s.t. } ||\boldsymbol{\mu}||_2 = 1, \quad (3)$$

where $\mathbf{A}$ is the Gram matrix of kernel $A$ on the training set, $\langle \mathbf{A}, \mathbf{B}\rangle = tr(\mathbf{A}\mathbf{B}^T)$ and $||\mathbf{A}||_F^2 = tr(\mathbf{A}\mathbf{A}^T)$.

The above optimization problem can be re-written in our terminology of K-examples as follows:

$$\max_{\boldsymbol{\mu}\ge 0} \frac{\boldsymbol{\mu}^T \left(\sum_{t_{ij}=1} \mathbf{z}_{ij} - \sum_{t_{ij}=-1} \mathbf{z}_{ij}\right)}{\sqrt{\boldsymbol{\mu}^T \left(\sum_{\forall i,j} \mathbf{z}_{ij}\mathbf{z}_{ij}^T\right)\boldsymbol{\mu}}}, \text{s.t. } ||\boldsymbol{\mu}||_2 = 1$$

When the base kernels are centered, as proposed in (Cortes et al., 2010b), the denominator represents the overall standard deviation of the projections of the K-examples on the vector $\boldsymbol{\mu}$. Hence target alignment attempts to find a projection direction $\boldsymbol{\mu}$ that maximize the difference between the sums of the projections of the positive and negative K-examples, while minimizing the overall variance of the projected K-examples. This is very similar to using Fisher-LDA in the K-space, with non-negativity constraints on $\boldsymbol{\mu}$. In fact, viewing target alignment from this perspective, makes it clear that it implicitly makes the assumption that the data is homoscedastic (the positive and negative K-examples have the same covariance), which might not be appropriate in real applications.

### 2.2. Connection to Learning with Hyperkernels

The approach proposed in this paper can also be cast in the framework of learning with hyperkernels (Ong et al., 2005) which provides a general recipe for kernel learning and includes Multiple Kernel Learning as a special case. It introduces the notions of kernel *quality functional*, a measure of "goodness" of a kernel that depends on the training data, and *Hyper Reproducing Kernel Hilbert Space*, an RKHS over kernel functions that defines the class of kernels that can be learned. Once the desired Hyper-RKHS and quality functional are specified, one has to solve a semi-definite program (SDP) to optimize the quality functional regularized by the norm induced by the Hyper-RKHS.

When using an SVM as the K-classifier, TS-MKL can be put in the learning with hyperkernels framework by defining the Hyper-RKHS to be the set of non-negative linear combinations of base kernels, and the quality functional to be the hinge loss in K-space. Considering this specific setting has significant advantages: it enables the use of simple and well understood binary classification techniques to learn the kernel, it enables a theoretical analysis, and it allows a significantly more scalable implementation. Equally important, all these advantages do not seem to come at the cost of reduced performance, as we are still performing on par with or better than competing MKL techniques.

## 3. Theoretical Results[2]

In this section we make the connection between the performance a K-classifier in the K-space and the performance on the original problem precise. This justifies the approach taken in this paper not only intuitively, but also from a theoretical standpoint. Specifically, we bound the generalization error of an SVM that uses the kernel induced by a K-classifier in terms of the expected hinge loss and the margin of the K-classifier in the K-space:

**Theorem 3.1** *Let $P$ be a distribution on $\mathcal{X} \times \{\pm 1\}$, $z_{xx'}$ and $t_{yy'}$ be as in Equation 1, $h$ be a K-classifier, and $R$ be a constant s.t. $h(z_{xx}) \leq R^2$ $\forall x \in \mathcal{X}$. Let*

$$HL_{h,\gamma} = E_{((x,y),(x',y'))\in P\times P}\left[\!\!\left[\left[1 - \frac{t_{yy'}h(z_{xx'})}{\gamma}\right]_+\right]\!\!\right]$$

*be the expected K-space hinge loss relative to margin $\gamma$ of the K-classifier $h$. Then, with probability $1-\delta$, a classifier $\hat{f}$ with generalization error*

$$P_{(x,y)}\left[\!\!\left[y\hat{f}(x) \leq 0\right]\!\!\right] \leq HL_{h,\gamma} + \mathcal{O}\left(\sqrt{\frac{R^4\ln(1/\delta)}{\gamma^2 n}}\right)$$

*can be learned efficiently from a training sample of $n$ instances drawn IID from $P$.*

[2]Due to lack of space, all proofs are included in the supplementary material.

The theorem follows from the two lemmas stated below. The first lemma shows that a K-classifier that has low expected hinge loss in the K-space will induce a "good" kernel. The second lemma shows that a good kernel allows for a classifier with low generalization error to be efficiently learned from a finite training sample. The following definition states formally what we mean by a good kernel (Srebro, 2007).[3]

**Definition** A kernel $K$ is an $(\epsilon,\gamma)$ good kernel in hinge loss with respect to a distribution $P$ on $\mathcal{X}\times\{\pm 1\}$ if there exist a classifier $w \in \mathcal{H}_K$ with $\|w\|_{\mathcal{H}_K} = 1$ s.t.

$$E_{(x,y)}\left[\!\!\left[\left[1 - \frac{y\langle w, \phi(x)\rangle}{\gamma}\right]_+\right]\!\!\right] \leq \epsilon$$

where $\mathcal{H}_K$ is the Hilbert space and $\phi(\cdot)$ is the feature mapping corresponding to $K$.

**Lemma 3.2** *Let $P$, $h$, $HL_{h,\gamma}$, $R$ be as in Theorem 3.1. Then the $\tilde{K}_h$ is a $(HL_{h,\gamma}, \frac{\gamma}{R})$ good kernel in hinge loss with respect to $P$.*

**Lemma 3.3** *Let $K$ be an $(\epsilon,\gamma)$ good kernel in hinge loss, with $K(x,x) \leq R^2$ $\forall x \in \mathcal{X}$ . Let $(x_i, y_i)_{i=1}^n$ be an IID training sample, and $\hat{f}(x) = \hat{w}\cdot\phi(x)$ with*

$$\hat{w} = \arg\min_{||w||_{\mathcal{H}_K}\leq 1} \frac{1}{n}\sum_{i=1}^{n}\left[1 - \frac{y_i w\cdot\phi(x_i)}{\gamma}\right]_+$$

*be a kernel classifier that minimizes the average hinge loss relative to $\gamma$ on the training sample. Then, with probability at least $1-\delta$, we have:*

$$P_{(x,y)}\left[\!\!\left[y\hat{f}(x) \leq 0\right]\!\!\right] \leq \epsilon + \mathcal{O}\left(\sqrt{\frac{R^2\ln(1/\delta)}{\gamma^2 n}}\right)$$

Lemma 3.3 follows directly from Theorem 21 in (Bartlett & Mendelson, 2002).

Note that, unlike in the one-stage kernel learning case, the generalization bound in Theorem 3.1 is in terms of the expected hinge loss of the K-classifier not the training hinge loss. While we are hopeful a generalization bound for the classification problem in the K-space can be obtained, as of now it remains an open problem.

We can, however, prove a concentration bound for the expected hinge loss of a K-classifier. This is the analog of the concentration bounds for target alignment in (Cortes et al., 2010b; Cristianini et al., 2001).[4]

[3]A kernel that does not satisfy this definition is not necessarily a "bad" kernel. We just can not make any formal statements with respect to its performance.

[4]This is not a regular generalization bound as the K-classifier is not allowed to depend on the IID sample.

**Theorem 3.4** *Let $P$, $h$, $HL_{h,\gamma}$, $R$ be as in Theorem 3.1. Let $(x_i, y_i)_{i=1}^n$ be an IID sample distributed according to $P$. Then the following inequality holds with probability at least $1 - \delta$*

$$HL_{h,\gamma} \leq \frac{2}{n(n-1)} \sum_{1 \leq i < j \leq n} \left[ 1 - \frac{t_{ij} h(z_{ij})}{\gamma} \right]_+ + \sqrt{\frac{2\left(1 + \frac{R^2}{\gamma}\right)^2 \ln 1/\delta}{n}}$$

## 4. Empirical Evaluation

We evaluate the proposed method on two object recognition datasets (Caltech-101 and Caltech-256), three bioinformatics datasets (Psort+, Psort- and Plant), and four UCI datasets (Sonar, Pima, Vertebral and Ionosphere). We compare our method with several baselines: best kernel, uniform combination of base kernels (Average), target alignment, and the one-stage MKL algorithms SILP (Sonnenburg et al., 2006), SimpleMKL (Rakotomamonjy et al., 2007), $L_2$-Norm MLK (Kloft et al., 2011), and UFO-MKL (Orabona & Jie, 2011). For two-stage methods we use LIBSVM (Chang & Lin, 2011) to train the data classifier and select the regularization parameter $C$ via 4-fold cross-validation for all datasets except Caltech where it is fixed at 1000. On multi-class problems, we use a one-vs-rest SVM. For one-stage approaches other than UFO-MKL, we selected $C$ as above and use a one-vs-rest scheme for multi-class problems. For UFO-MKL we use the joint multi-class formulation and search over $\alpha$ and $C$ using a bi-dimensional grid. Following (Orabona & Jie, 2011), we run the optimization for 20 epochs on UCI datasets, 30 epochs on Caltech-101 and 100 epochs on Caltech-256. All kernels used in the experiments are centered and standardized to have zero mean and unit variance in feature space.

### 4.1. Methodology for TS-MKL

To learn kernel combination weights $\boldsymbol{\mu}$ with TS-MKL we optimize the objective in Eq. 2 using Pegasos (Shalev-Shwartz et al., 2007) with an additional projection to the non-negative constraint set after each sub-gradient step. We use a batch size of 100 for each sub-gradient computation and run $10^3$ sub-gradient steps for UCI datasets and $10^5$ for all others. Figure 2, plots the test data accuracy versus the number of gradient iterations on Caltech-101, showing that after $10^5$ iterations the change in accuracy is minimal. For the bigger Caltech-256 there is also essentially no change after $10^5$ iterations. We use subsampling to balance the positive and negative K-examples.

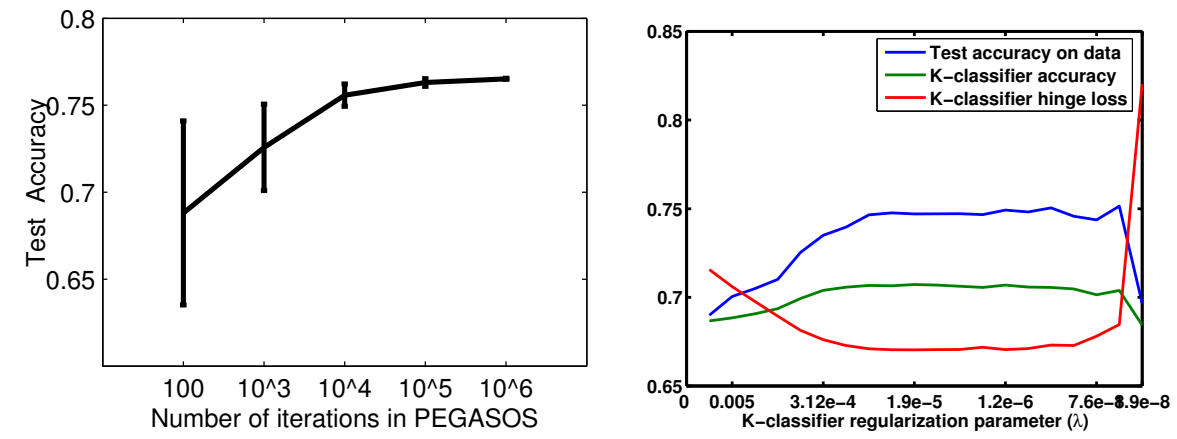

*Figure 2.* **Left:** Test data accuracy as a function of number of sub-gradient iterations in Pegasos. **Right:** Correlation between hinge loss (and accuracy) on K-examples and test data accuracy on Caltech-101.

To select the parameter $\lambda$, we use a single 80%-20% random split of the Pegasos training set and search for the $\lambda$ with the lowest validation hinge loss[5]. The search grid for $\lambda$ is taken to be in the range of 100 to $10^{-8}$ dividing in each step by 4. A big advantage of this selection scheme for $\lambda$ is that it is completely independent from the data classifier that will ultimately use the learned kernel. This keeps the setup simple and avoids intricate multi-level multi-dimensional validation schemes across the parameters of the data classifier and the K-classifier. Fig. 2, shows the hinge-loss in K-space, the accuracy of the K-classifier, and the accuracy of the data classifier that uses the learned kernel, as a function of $\lambda$. The plot shows a clear correlation between hinge loss in K-space and data accuracy with the learned kernel. The data accuracy increases when the hinge loss in K-space decreases and vice versa. This experiment provides further emprical evidence for our theoretical results that show that a good K-classifier (having low hinge loss in K-space) corresponds to a good learned kernel.

After $\lambda$ is selected, Pegasos is retrained on the full training set of K-examples. The obtained weight vector $\boldsymbol{\mu}$ is then used to linearly combine the base kernels, and the SVM data classifier is trained using this learned kernel with $C$ selected as described above.

### 4.2. Caltech-101 and Caltech-256

Both these datasets contain pictures of objects and the task is to recognize the object category. Caltech-101 has 102 classes and Caltech-256 has 256 classes. Caltech-101 is perceived as an easier dataset than Caltech-256 in which images are not left-right aligned and there are more categories. We follow the experimental setup used in (Gehler & Nowozin, 2009) and use the same 39 base kernels and train test splits.

We report results using all 102 classes for Caltech-

[5]Since the K-examples are dependent, the training and validation set will not be fully independent. Nevertheless, this does not seem to negatively affect the performance.

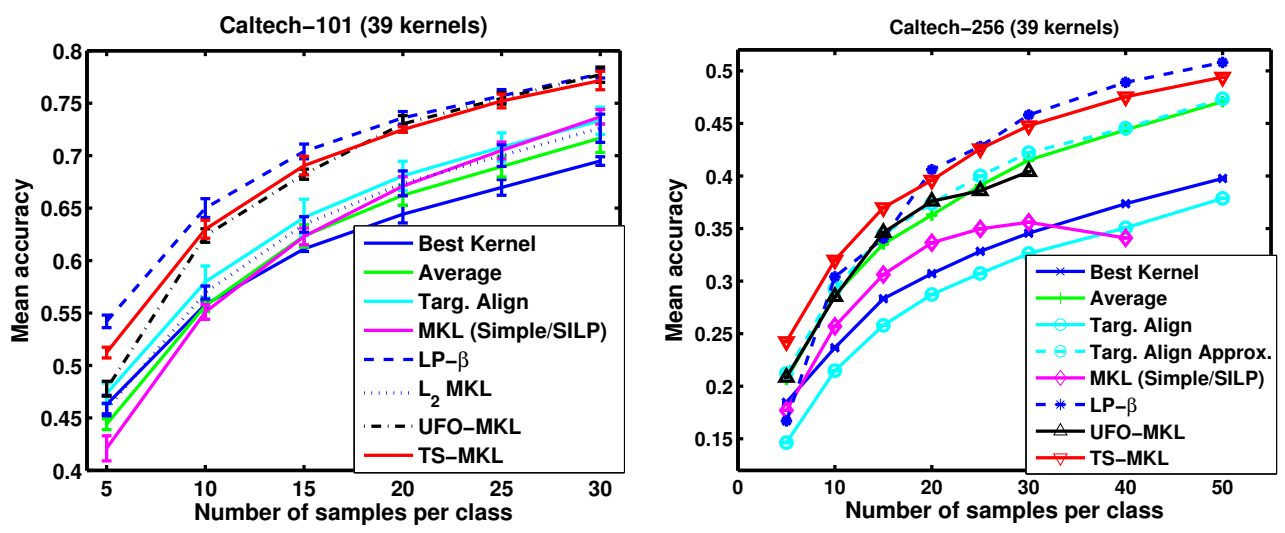

*Figure 3.* **Left:** Caltech-101 results: mean accuracy over all classes for different sample sizes, averaged over 5 splits. **Right:** Caltech-256 results: mean accuracy over all classes for different sample sizes

101 averaged over five splits. For Caltech-256, the results are for 256 classes (excluding the clutter category), for a single split. The performance measure used is mean prediction rate per class. The number of training images per class is varied in the range $5, 10, 15, 20, 25, 30$ for Caltech-101, and in the range $5, 10, 15, 20, 25, 30, 40, 50$ for Caltech-256. The number of test images used is up to 50 images per class for Caltech-101 and 25 images per class for Caltech-256. The regularization parameter for the data SVM, C, is fixed to 1000 for all methods[6].

The results for Caltech-101 and Caltech-256 are shown in Fig. 3.[7] On Caltech-101 our approach yields a mean accuracy of 0.512, 0.630, 0.691, 0.725, 0.752, 0.772 for 5, 10, 15, 20, 25, 30 samples per class respectively. Comparing to UFO-MKL, our performance is higher for 5 samples per class, and very similar for all other sample sizes. One-stage MKL methods using the one-vs-all multi-class scheme perform significantly worse and do not even outperform the average kernel until the training set has 25 samples per class. This is probably because data is too scarce to allow learning a separate kernel for each class. Target alignment performs a little better than the average kernel, but is still significantly worse than TS-MKL. We also show the performance of LP-$\beta$ (Gehler & Nowozin, 2009), which, to the best of our knowledge, is the state of the art method on this data set[8]. The performance of TS-MKL and UFO-MKL is almost on par with LP-$\beta$, especially for larger sample sizes. While LP-$\beta$ is similar in spirit to multiple-kernel learning, it is not a true kernel learning algorithm as it does not produce a kernel, but rather learns an ensemble of SVM classifiers, each of which is trained on an individual kernel.

On Caltech-256 dataset, our approach performs better than all competing kernel learning baselines. We achieve 0.245, 0.320, 0.370, 0.426, 0.448, 0.475, 0.494 mean accuracy for 5, 10, 15, 20, 25, 30, 40, 50 training samples per class. This performance is significantly higher than the best results reported in the literature for 5, 10, and 15 training samples, after which we again perform on par with LP-$\beta$. On this dataset, UFO-MKL performance[9] is similar to that of the average kernel, while the rest of the one-stage MKL techniques perform worse. Exact target alignment is worst among all other approaches, however approximate target alignment is able to at least match the performance of the average kernel.

### 4.3. Bioinformatics datasets

We evaluate our method on a problem relevant to cell-biology predicting: the subcellular localization of proteins, which is crucial in making inference about protein function and protein interactions. We follow the experimental setup of (Zien & Ong, 2007) and use the same 69 kernels. The kernels used are: 2 kernels on phylogenetic trees, 3 kernels from BLAST E-values and 64 sequence motif kernels.

We experiment with three datasets. The first two datasets are for the problem of bacterial protein locations (Gardy et al., 2004). The Psort+ dataset has 541 data points with 4 classes and Psort– dataset has 1444 data points with 5 classes. We report average F1 score over all classes over 10 random splits for both these datasets as done in (Zien & Ong, 2007). The third dataset used is the original plant dataset of TargetP (Emanuelsson et al., 2000), and has 940 examples with 4 classes. We use the performance measure of Matthew's Correlation Coefficient (MCC) following the evaluation in (Zien & Ong, 2007). Again, average MCC score over all 4 classes is reported.

The results are shown in Table 1. The papers that have used the Psort datasets in the past (Gardy et al., 2004; Zien & Ong, 2007), reported results after filtering out the most unsure predictions in the test set. For Psort+ and Psort-, about 15% and 13.3% of the test examples were filtered out respectively and the performance is reported only for the remaining predictions. We follow the same procedure to be able to compare with these methods. We also report performance for full test set. On these datasets, all the kernel learning methods have similar performance, and are better than the best kernel and average kernel baselines. Multi-class multiple kernel learning (MC-MKL) of (Zien & Ong, 2007) is also close to our method and other baselines.

[6] $C = 1000$ is the best setting for the one-stage MKL algorithms (Gehler & Nowozin, 2009)

[7] We take the results for LP-$\beta$ and MKL from (Gehler & Nowozin, 2009).

[8] LP-$\beta$ achieves state of the art performance when using additional kernels (`http://www.vision.ee.ethz.ch/~pgehler/projects/iccv09`). We could not obtain all 48 kernels, so we only report results with only 39 kernels for all methods

[9] The UFO-MKL performance at 40 and 50 samples is missing because the code we are using runs out of memory.

| | Psort+ | | Psort− | | Plant |
|---|---|---|---|---|---|
| | Full test | Filtered | Full test | Filtered | Full test |
| Best Kernel | 81.30(4.69) | 86.26(4.96) | 85.95(1.54) | 91.53(1.04) | 72.19(3.94) |
| Average | 84.75(3.97) | 89.48(4.97) | 88.03(1.10) | 93.95(1.14) | 86.72(3.38) |
| Target Alignment | 88.14(3.99) | 92.82(3.99) | 89.91(1.42) | 95.22(1.33) | 89.13(2.75) |
| MKL (SILP/Simple) | 89.05(3.02) | 93.89(3.37) | 91.01(1.10) | 96.01(1.51) | 89.32(2.76) |
| MC-MKL | – | 93.8 | – | 96.1 | 89.1 |
| TS-MKL(Our Approach) | 89.08(3.32) | 93.50(2.74) | 90.15(1.33) | 95.63(1.31) | 88.86(3.26) |

*Table 1.* Average accuracy measures (%) over 10 splits for Psort+, Psort- and Plant datasets. Numbers in parentheses are the std. deviations. The accuracy measures for MC-MKL (Zien & Ong, 2007) are taken from their paper.

### 4.4. UCI datasets

We use four UCI datasets: Sonar, Ionosphere, Pima and Vertebral (the three class version). For each of these datasets, we perform two types of MKL experiments. In first setting, we construct a total of 13 kernels on the full feature vectors: 9 Gaussian kernels ($e^{-\gamma||\mathbf{x}_i - \mathbf{x}_j||^2}$) with $\gamma = \{2^{-10}, 2^{-9}, \ldots, 2^{-2}\}$, 3 polynomial kernels of degree 2,3 and 4, and a linear kernel. In the second setting, we augment these 13 kernels with another set of Gaussian, polynomial and linear kernels constructed on individual features of the data. The range of parameter $\gamma$ for Gaussian and degree parameter for polynomial kernel is kept same as before. If the data has $d$ features, we have total $13d + 13$ kernels in the second setting. We report average accuracy accuracy over 10 random $80\% - 20\%$ train-test splits.

The results are shown in Table 2. On all these datasets, no kernel learning approach seems to improve performance over the straightforward baselines of best kernel and average kernel. Although further study is needed to reach a definite conclusion, these results seem to indicate that blindly using a kitchen sink of standard kernels is not beneficial if the goal is to combine these kernels using an MKL approach. This highlights the importance of evaluating MKL techniques using datasets like Caltech and PSORT, where the kernels have been carefully designed using domain knowledge to capture different, potentially useful, notions of similarity in the data.

### 4.5. Computational Efficiency

Since the number of K-examples is quadratic in the number of training instances, one might worry about the scalability of the TS-MKL method. In this section we compare the running time of TS-MKL with Target Alignment, and UFO-MKL (Ultra-Fast Optimization MKL) which, to the best of our knowledge, is the fastest one-stage MKL technique to date.

| | Sonar p = 793 | Pima p = 117 | Caltech 101 Train 30 |
|---|---|---|---|
| Targ. Align | 133 | 93.71 | 607(579) |
| UFO-MKL | 3.018 | 17.97 | 387 |
| TS-MKL | 1.09 | 1.3977 | 34(6) |

*Table 3.* Running time in seconds. In paranthesis we show the time taken by the kernel learning stage alone.

Table 3 shows the running times for the Sonar, Pima and Caltech 101 datasets. The running time is for a single run using the best setting of parameters (i.e. it does not include the time for parameter selection). For TS-MKL and Target Alingment we also show in paranthesis the time taken by the kernel learning stage alone, without the final data SVM, on Caltech-101. For Sonar, which has only 166 training samples, the running time of UFO-MKL and TS-MKL is comparable. However, on Pima, which has 614 samples, and on Caltech, which has 3060 samples and 102 classes, TS-MKL is more than an order of magnitude faster than UFO-MKL. This shows that, by taking advantage of the advances in large scale stochastic optimization, TS-MKL is not only able to gracefully handle the quadratic increase in the number of K-examples, but it is actually the fastest MKL method to date.

## 5. Conclusions and Future Work

Framing kernel learning as a standard classification problem in a properly defined instance space allows us to easilly adapt well understood classification techniques to obtain a scalable and high performing two-stage multiple kernel learning algorithm. Our approach is backed up by formal theoretical guarantees, and by empirical evaluation that shows it always outperforms or is on par with leading one-stage and two-stage kernel learning methods. This is a remarkable feat for a method that is quite simple and intuitive.

This new perspective on multiple kernel learning opens the door to a number of interesting questions to be addressed in subsequent research. Examples are: exploring the use of non-linear K-classifiers in conjunction with the learning with similarity functions framework; improving performance in scarce data condi-tions through semi-supervised and multi-task multiple kernel learning by using such techniques to learn the K-classifier; or applying TS-MKL to semi-supervised clustering and dimensionality reduction problems where the supervised signal is usually given in terms of pairwise must-link and can-not-link constraints rather than labels.

| | Sonar | | Ionosphere | | Pima | | Vertebral | |
|---|---|---|---|---|---|---|---|---|
| | p = 793 | p = 13 | p = 442 | p = 13 | p = 117 | p = 13 | p = 91 | p = 13 |
| Best Kernel | 86.90(4.23) | 86.90(4.24) | 95.00(2.04) | 95.00(2.04) | 76.10(2.63) | 76.10(2.63) | 83.65(5.73) | 83.67(5.73) |
| Average | 85.00(4.5) | 86.42(3.73) | 92.00(2.87) | 94.28(3.01) | 76.82(2.76) | 76.30(2.62) | 81.58(5.92) | 82.03(5.03) |
| Targ. Align | 80.24(4.2) | 85.47(3.26) | 91.57(2.28) | 94.42(2.17) | 75.97(3.03) | 76.82(3.15) | 82.88(6.18) | 80.90(4.18) |
| MKL(SILP/Simple) | 85.23(5.11) | 84.76(2.55) | 92.54(1.56) | 95.42(2.50) | 75.71(3.28) | 75.97(3.16) | 82.72(4.16) | 78.42(3.55) |
| $L_2$-MKL | 86.42(4.05) | 85.71(4.04) | 91.85(1.51) | 95.14(2.04) | 75.45(2.31) | 76.55(2.23) | 79.68(4.84) | 80.87(5.1) |
| UFO-MKL | 82.85(6.7) | 86.19(4.3) | 91.85(2.86) | 96.14(1.9) | 74.28(2.47) | 74.02(3.46) | 79.16(6.57) | 79.09(6.03) |
| TS-MKL(Our Approach) | 86.43(3.9) | 86.19(3.38) | 92.43(1.18) | 94.29(2.12) | 75.78(3.02) | 76.42(2.87) | 82.82(5.63) | 81.10(4.42) |

*Table 2.* Average accuracy (%) over 10 random splits on UCI datasets. $p$ denotes the number of base kernels. Numbers in parentheses are the std. deviations.